\documentclass[journal]{IEEEtran}

\ifCLASSINFOpdf
  \usepackage[pdftex]{graphicx}
\else
  \usepackage[dvips]{graphicx}
\fi
\usepackage{amsmath}
\usepackage{amsfonts}
\usepackage{amssymb}
\usepackage{amsthm}
\usepackage{algorithmic}
\usepackage{algorithm}
\usepackage[export]{adjustbox}
\usepackage{booktabs}

\usepackage{array}
\usepackage{tabularx}
\usepackage{multirow}
\usepackage{rotating}
\usepackage{array}

\usepackage[caption=false,font=footnotesize]{subfig}

\usepackage{changepage}

\usepackage{url}
\usepackage{siunitx}
\begin{document}

\title{Application of Gated Recurrent Units for CT Trajectory Optimization}
\pagenumbering{gobble}

\author{Yuedong~Yuan\thanks{Y. Yuan, L. Schneider and A. Maier are with the Pattern Recognition Lab, Friedrich-Alexander University Erlangen-Nuremberg, Erlangen, Germany (e-mail:yuedong.yuan@fau.de)},~Linda-Sophie~Schneider, Andreas~Maier,~\IEEEmembership{Member,~IEEE} }

\maketitle

\begin{abstract}
Recent advances in computed tomography (CT) imaging, especially with dual-robot systems, have introduced new challenges for scan trajectory optimization. This paper presents a novel approach using Gated Recurrent Units (GRUs) to optimize CT scan trajectories. Our approach exploits the flexibility of robotic CT systems to select projections that enhance image quality by improving resolution and contrast while reducing scan time. We focus on cone-beam CT and employ several projection-based metrics, including absorption, pixel intensities, contrast-to-noise ratio, and data completeness. The GRU network aims to minimize data redundancy and maximize completeness with a limited number of projections. We validate our method using simulated data of a test specimen, focusing on a specific voxel of interest. The results show that the GRU-optimized scan trajectories can outperform traditional circular CT trajectories in terms of image quality metrics. For the used specimen, SSIM improves from 0.38 to 0.49 and CNR increases from 6.97 to 9.08. This finding suggests that the application of GRU in CT scan trajectory optimization can lead to more efficient, cost-effective, and high-quality imaging solutions.

\end{abstract}


%
\IEEEpeerreviewmaketitle

\section{Introduction}

\IEEEPARstart{C}{omputed} Tomography (CT) has transformed non-destructive testing by providing unprecedented insight into the structure of objects. Recent advances in CT technology have introduced dual-robot CT systems, where individual robots control the X-ray source and detector. This innovation provides flexibility by allowing projections from arbitrary views around the object of interest. Robotic CT systems offer the potential for more customized and complex CT trajectories, facilitating scan trajectory optimization \cite{herl2020scanning}. Optimized scan trajectories reduce scan time and radiation exposure while maintaining or improving image quality. This increases scanner efficiency, reduces operating costs, and increases throughput. Furthermore, reconstruction through carefully selected projections can improve image resolution and contrast, which are critical for detecting finer details and subtle anomalies in scanned objects \cite{schneider2022learning}.

This paper presents a new approach to CT scan trajectory optimization using Gated Recurrent Units (GRUs), a machine learning technique, based on several projection-based metrics, including absorption, pixel intensities, contrast-to-noise ratio (CNR) \cite{desai2010practical}, and data completeness \cite{herl2022xray}. Our methodology evaluates and selects projections for data reconstruction using these metrics. This study uses cone-beam CT, focusing on a specific volume of interest (VOI) within the scanned object. We present GRU as a powerful tool for optimizing CT scan trajectories. The approach aims to improve the accuracy and efficiency of CT scanning processes, paving the way for more cost-effective and higher-quality imaging. Our results show that the GRU-optimized scan trajectories are superior to traditional circular CT trajectories in terms of image quality, indicating potential advances in CT imaging techniques.

\section{Methods}

\subsection{Absorption and Pixel Intensity} \label{sec: pixel intensity}
In X-ray imaging, beam hardening and scattering have a significant impact on image quality, often resulting in artefacts in CT images \cite{pettersson2021comparison}. These problems arise from the energy-dependent attenuation of X-rays, which is influenced by the composition of the material and the different depths at which they penetrate. The phenomenon of beam hardening occurs when low energy X-rays are absorbed more strongly than their high energy counterparts as they pass through an object \cite{barrett2004artifacts}. This differential absorption changes the energy profile of the beam. To quantify this effect, we use a metric known as pixel intensity, representing the residual intensity of X-rays captured by the detector. This residual intensity of X-rays is reflected in the pixel values of the projection images. The value of 70-quantiles of pixel values within the region of interest (ROI) is calculated to evaluate this metric. A higher pixel intensity value indicates lower attenuation, which is desirable for optimizing CT scan trajectories. To mitigate the effects of beam hardening, projections with a pixel value within the ROI below a certain threshold $\alpha$ are excluded from the analysis. It is important to note that this threshold varies depending on the specific object being scanned and it is an empirical value.

\subsection{Contrast-to-Noise Ratio}
CNR, a core image quality metric \cite{desai2010practical}, helps evaluate the detectability of defects in projection images \cite{tokunagasimulation}. The CNR is defined by the equation below.
\begin{equation}
CNR = \frac{{\left| {Max - Min} \right|}}{{{\sigma _n}}}
\label{eq:CNR}
\end{equation}
where $Max$ and $Min$ are the highest and lowest pixel values within the ROI, respectively, and ${\sigma _n}$ is the standard deviation of the background pixels.

A higher CNR means better contrast differentiation within the ROI, indicating better image quality and projection clarity.

\subsection{Data Completeness}
The data completeness metric, rooted in Tuy's Data Sufficiency Condition, evaluates the scan trajectory of CT systems for accurate volumetric reconstruction of the VOI within scanned objects. Tuy's Data Sufficiency Condition necessitates that each cross-sectional plane of the scanned object intersects the scan trajectory at least once in order to reconstruct the scanned object \cite{tuy1983inversion}. A single circular trajectory, for instance, fails to satisfy this condition, since there is no intersection between the single circular trajectory and the cross-sectional planes parallel to this trajectory plane or slightly tilted to this trajectory plane. 

\begin{figure}[!t]
\centering
\includegraphics[width={0.8\linewidth}]{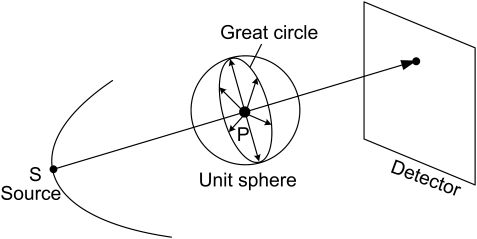}
\caption{A great circle formed by normal vectors of all planes containing the X-ray path $SP$ \cite{liu2012completeness}.}
\label{fig:great circle}
\end{figure}

Figure \ref{fig:great circle} illustrates the terms great circle and unit sphere. Each sample point on the unit sphere represents a normal vector of the cross-sectional plane passing through a scanned object point within the VOI, known as the Radon plane. In addition, each scan trajectory point corresponds to a great circle. The plane of the great circle is perpendicular to the viewing direction of the X-ray beam from the source location $S$ to the scanned object point $P$, for which the data completeness is computed.

To achieve a discrete formulation of the data completeness metric, we build on the concepts from \cite{maier2015discrete} and their extensions in \cite{herl2022xray}. The procedure involves two main steps. First, we sample $M$ vectors $u$ from the unit sphere as uniformly as possible. Note that only half of the sphere needs to be sampled due to the symmetry of the plane integrals in the upper and lower halves. Second, we assess the unit sphere coverage for each voxel in the VOI. This involves identifying unit sphere vectors $u$ that align with the normal vectors of the Radon planes. Only viewing vectors $d$ that hit the detector are considered.

The Nyquist-Shannon sampling theorem informs the requirement that the angular distance between adjacent planes in Radon space must be less than a maximum gap $\Delta \gamma$. Given that the inner product of the two unit vectors represents the cosine of their angular separation, vectors of interest meet the criterion below.
\begin{equation}
|d^Tu| < \sin ( \Delta \gamma )
\label{hit_check}
\end{equation}

This equation is evaluated for each unit vector $u$ against all viewing directions through the VOI, leveraging the constant nature of $\Delta \gamma$ for precomputed sine function values. The complete summary of the steps for calculating the data completeness metric is described in Algorithm \ref{alg:completeness number}.
\begin{algorithm}
\caption{Calculation of data completeness matrix for $\mathbf{P}$}
\begin{algorithmic}
\label{alg:completeness number}
\STATE { Source positions $ S \in \mathbf{R}^{3 \times N}$}
\STATE { Unit vectors $ \mathbb{S}_P \in \mathbf{R}^{3 \times M}$}
\STATE {Initialize data completeness matrix $C \in {0,1}^{N \times M}$}
\FOR{$s_i \in S$}
    \STATE { Calculate viewing direction $ {\mathbf{d}}:=\frac{\mathbf{P}-\mathbf{s_i}}{\parallel \mathbf{P}-\mathbf{s_i} {{\parallel }_{2}}}$}
    \FOR{$u_j \in \mathbb{S}_P$}
        \IF {$|d^Tu| < \sin ( \Delta \gamma ) $ }  
          \STATE $C_{i,j} \gets 1 $
        \ELSE
            \STATE $C_{i,j}  \gets 0 $
        \ENDIF 
    \ENDFOR
\ENDFOR
\end{algorithmic}
\end{algorithm}

The data completeness matrix $C$ can be calculated before CT trajectory optimization. This matrix is boolean, with an entry $C_{i,j}=1$ indicating that the viewing direction associated with the source point $s_i$ falls within the $\Delta \gamma$ threshold for being perpendicular to the unit vector $u_j$.

As depicted in Figure \ref{fig:great circle}, the more X-ray source locations are employed as CT trajectory points, the more unit sphere points could be hit by one or more great circles. For a subset $S_{opt} \subseteq S$, the coverage $ c $ of the unit sphere and thus the region-based data completeness of the optimized CT trajectory $S_{opt}$ can be calculated by counting the columns having at least one non-zero elements for the sampled unit sphere $ \mathbb{S}_P$.

\subsection{Gated Recurrent Unit for CT Trajectory Optimization}
In the context of data completeness, redundant data occurs when the same point on the unit sphere is covered multiple times by different projections. CT trajectory optimization aims to minimize this redundancy and maximize the coverage of the unit sphere using a given number of projections. This task essentially translates into identifying an optimized combination of projections.

During each iteration of the optimization process, the goal is to select a projection that contributes the most to the unit sphere $\mathbb{S}_P$, considering the influence of previously selected projections. This makes the problem sequential, for which recurrent neural networks (RNNs) are appropriate. RNNs are able to learn from sequential data due to their internal memory and can predict subsequent sequences, making them well suited for this application \cite{fujii2021two}.

\begin{figure}[!t]
     \centering
     \begin{subfloat}
         \centering
         \includegraphics[width={0.8\linewidth}]{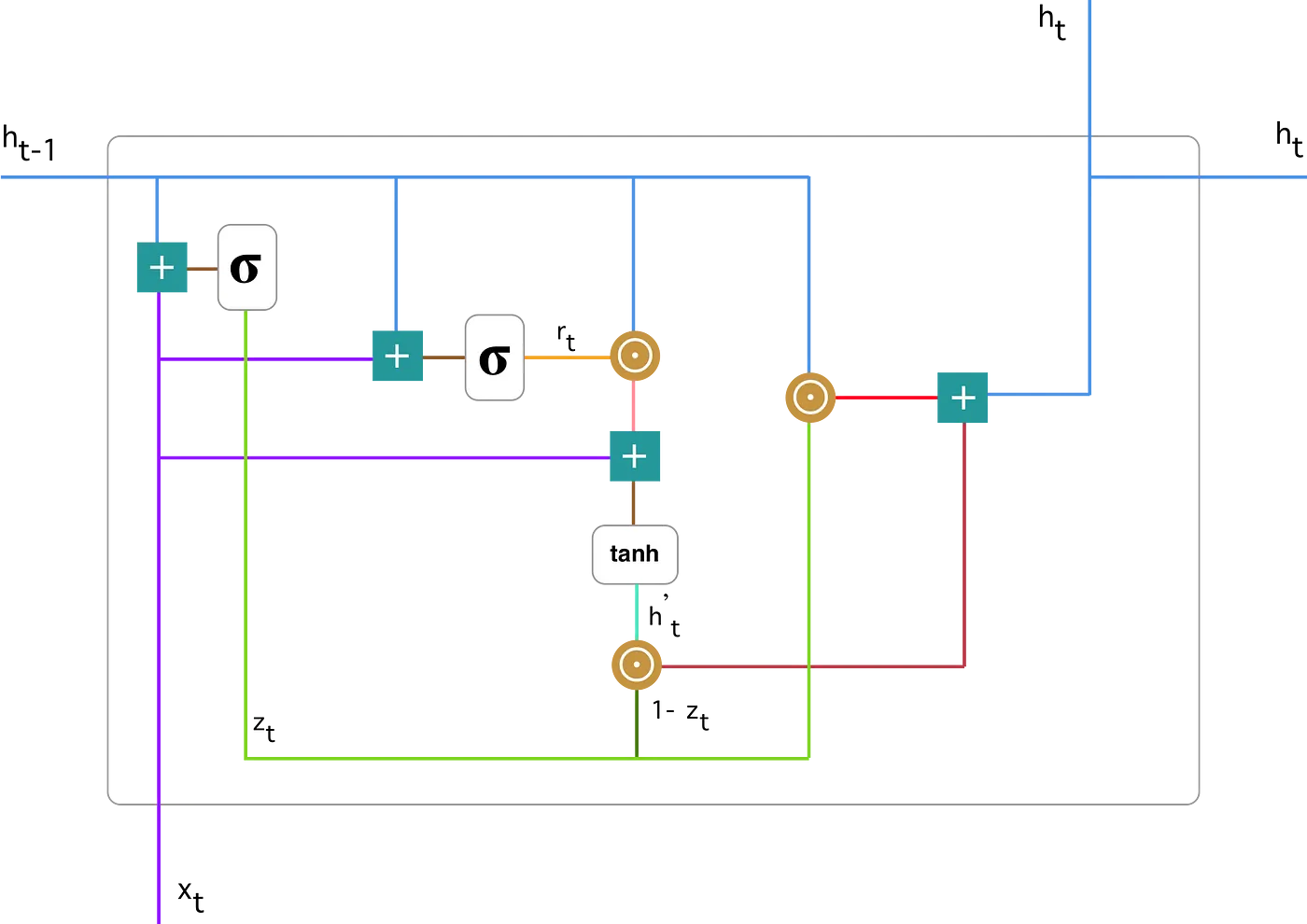}
     \end{subfloat}
     \hfill
     \begin{subfloat}
         \centering
         \includegraphics[width={0.9\linewidth}]{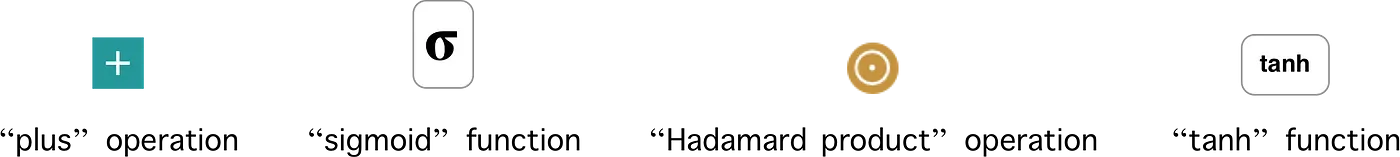}
     \end{subfloat}
    \caption{Scheme of a single GRU unit. ${x_t}$ denotes the input data at time step $t$, ${r_t}$ represents the reset gate at time step $t$, and ${z_t}$ corresponds to the update gate at time step $t$. ${h_t}_{ - 1}/{h_t}$ indicates the hidden state at time step $t-1/t$, while ${{h'}_t}$ represents the current memory of the unit at time step $t$ \cite{kostadinov2017understanding}.}
    \label{fig:GRU unit}
\end{figure}

A GRU is a recurrent neural network that has become popular for its computational efficiency and effectiveness in sequence-based tasks. As a streamlined alternative to the Long Short-Term Memory (LSTM) network, GRUs were designed to mitigate some of the computational challenges of LSTMs \cite{cho2014learning}. The distinctive feature of GRUs is their two-gate structure: the update gate and the reset gate, as opposed to the three gates of LSTMs. Figure \ref{fig:GRU unit} illustrates a single GRU unit. 

The update gate in GRUs controls the flow of information from previous steps to the current step, ensuring that only relevant data is passed on. Meanwhile, the reset gate decides how much past information is forgotten and how much new information is incorporated into the current step. This structure allows GRUs to model long-term dependencies in sequences efficiently. In addition, GRUs have no separate cell state, a simplification that often makes them easier to train, especially with smaller datasets. Given these advantages, GRUs are used in this research to optimize CT scan trajectories. Note that the multilayer GRU network is a sequence of copies of the same GRU unit, and the same units are used for all time steps. 

In CT trajectory optimization, pre-computed metrics serve as input to the GRU network. For each projection, these metrics include pixel intensity, CNR, and $M$ numbers for data completeness. Before providing the array of metrics to the network, specific projections are excluded, where the smallest pixel value in the ROI falls below the $\alpha$ threshold.

\begin{figure}[!t]
\centering
\includegraphics[width={1.0\linewidth}]{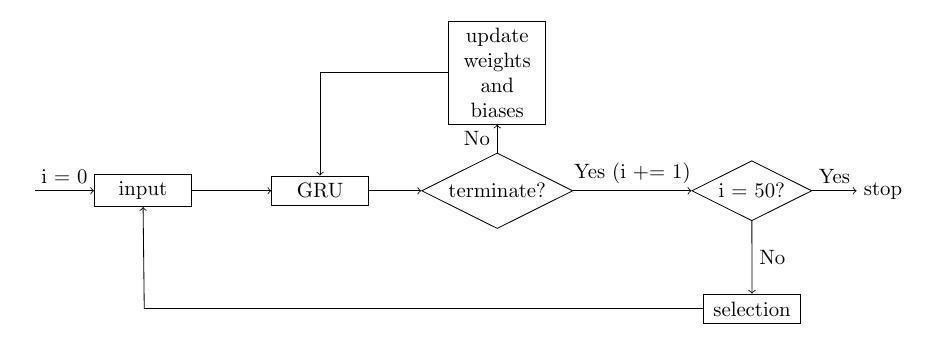}
\caption{Diagram of the selection process for the next projection through the GRU network} 
\label{fig:GRU training}
\end{figure}

Figure \ref{fig:GRU training} illustrates the iteration and projection selection process within the GRU network. During each iteration, the weights and biases of the GRU network are updated for a predefined number of training loops. A termination condition is then evaluated, and if it is met, the network selects a projection for that iteration. If the condition is not met, the network continues to be updated. Note that our network does not have a separate validation stage used commonly in traditional neural networks and the GRU network outputs a probability prediction for each batch. 

The loss function for the GRU network is defined as the binary cross entropy between the predicted probabilities and the scores of each batch. These scores are calculated using a weighted average of three normalized metrics: pixel intensity, CNR, and data completeness. Normalizing each metric to a scale of 0 to 1 mitigates the effects of their different value ranges. To emphasize data completeness, a weight factor of 16 is assigned. During the first iteration, the sequence in each batch contains a single element, i.e. a row of the input array. Thus, each one-element sequence contains only one projection. The training process stops when the loss in the current loop exceeds the loss in the previous loop, indicating that the loss value no longer decreases.

\begin{figure}[!t]
\centering
\includegraphics[width={1.0\linewidth}]{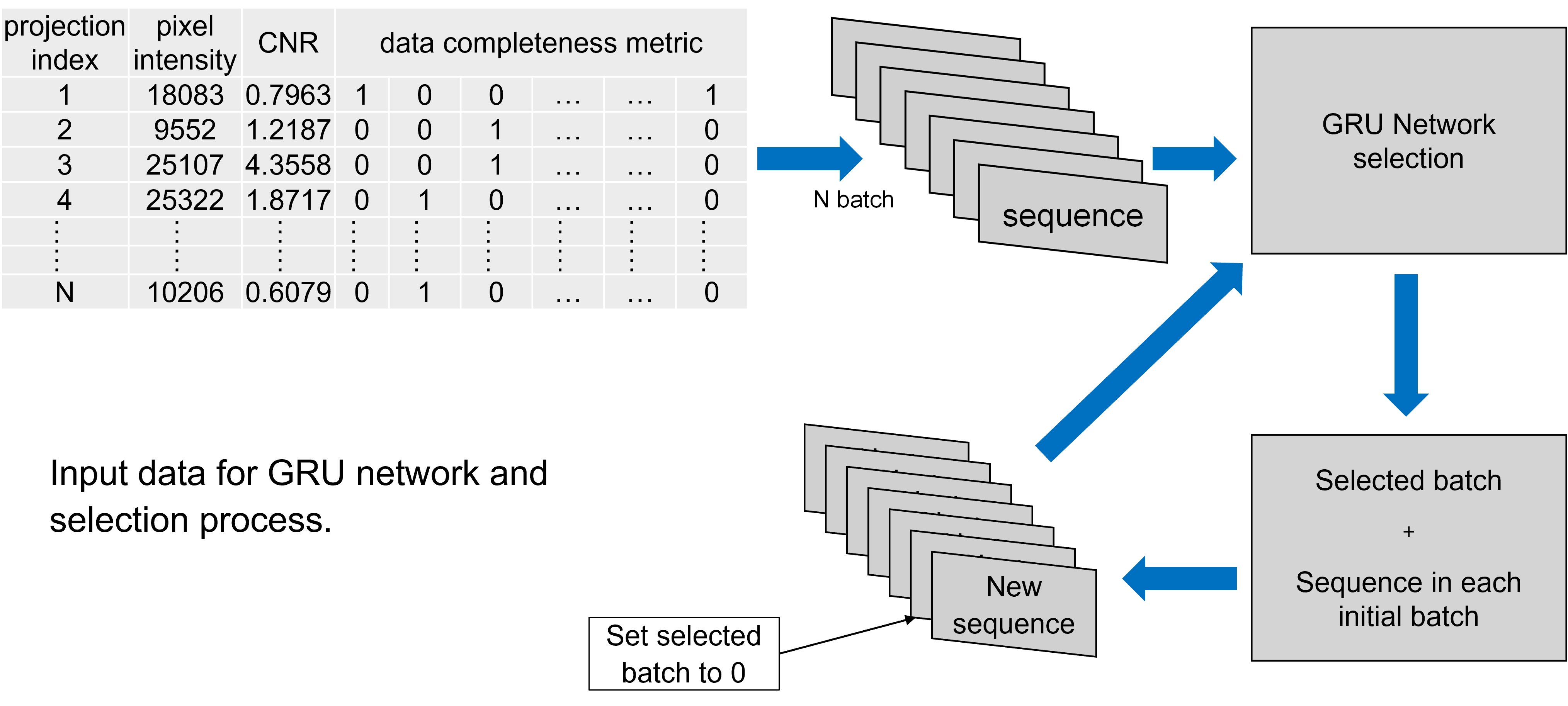}
\caption{Illustration of the data processing in GRU network.} 
\label{fig:data processing}
\end{figure}

After each selection, the data of the selected batch is added to each initial batch. This combination, including the chosen batch and the metrics of an available projection, forms the new input data for the next iteration. As the process is repeated, the sequence in each batch becomes progressively longer. To prevent the network from selecting the same projection twice, batches with an element already selected are assigned a value of 0. Figure \ref{fig:data processing} illustrates these data processing steps.

\section{Experiments and Results}

\begin{figure}[!t]
\centering
\subfloat[]{\includegraphics[height=1.2in]{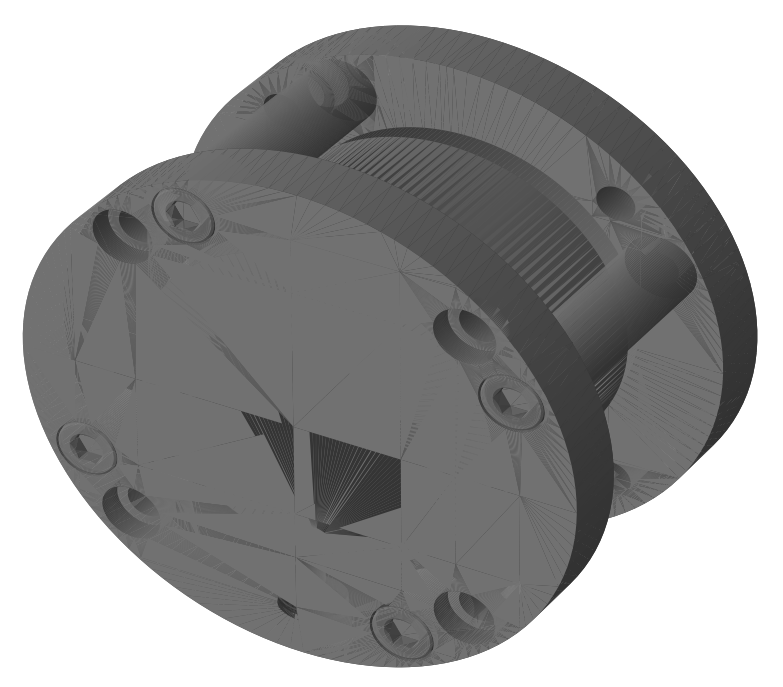}}
\hfil
\subfloat[]{\includegraphics[height=1.2in]{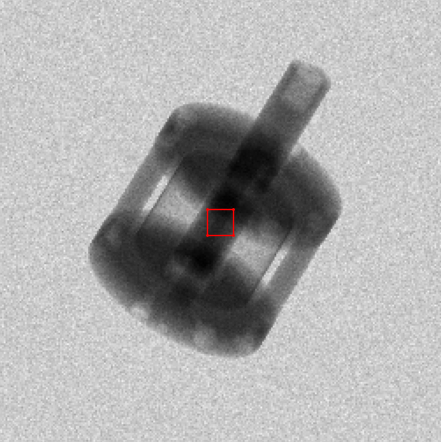}}
\caption{Test specimen overview. (a) Carbon specimen illustration. (b) Simulated projection highlighting the VOI in red.}
\label{fig:experiment_setup_2}
\end{figure}

We validated our CT scan trajectory optimization approach using a test specimen as shown in Figure \ref{fig:experiment_setup_2}, focusing on the center point as the VOI. For this test specimen, the center is complex to reconstruct completely because the X-rays' attenuation of some view directions is significantly strong due to the long transmission length.

We simulated 1000 spherical trajectory projections, filtering out those with high attenuation, reducing the number to 657. The simulation used a polychromatic spectrum corresponding to a tube voltage of \SI{150}{\kilo\volt}. A square detector of size \SI{11.52}{\centi\meter} and isotropic pixel spacing of \SI{450}{\micro\meter} is modeled. The chosen setup leads to a magnification of 2. Noise is simulated, but scatter effects are excluded from the simulation.

We sampled a unit sphere around our VOI, resulting in $\lvert \mathbb{S}_P \rvert = 1000$ unit sphere points and set a maximum angular distance of $\Delta \gamma = 0.573^{\circ} $.

Our GRU network is trained on a Quadro RTX 8000 GPU and the AdamW algorithm in PyTorch is used to update the weights. Using the Weights \& Biases platform, we have explored different hyperparameters for the GRU network to optimize the selection for a set of 50 projections. The final hyperparameters are presented in table \ref{tab: hyperparameter}.

\begin{table}[htbp]
\centering
\caption{Overview of optimized hyperparameters for GRU}
\begin{tabular}{@{}ccccc@{}}
\toprule
Parameter & hidden size & number of layers & training loop & learning rate \\
& & & & \\
\midrule
value & 1075 & 6 & 34 & 0.002677 \\
\bottomrule
\end{tabular}
\label{tab: hyperparameter}
\end{table}

\begin{figure}[thb]
		\begin{tabular}{r m{2.cm} m{2.cm} m{2.cm}} 
			\multicolumn{1}{c}{} & \multicolumn{1}{c}{\centering xy} & \multicolumn{1}{c}{\centering xz} & \multicolumn{1}{c}{\centering yz} \\	
			\multirow{2}{*}{\rotatebox[origin=c]{90}{Reference}} & \adjustbox{valign=m}{\includegraphics[width=2.3cm]{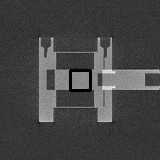}} & \adjustbox{valign=m}{\includegraphics[width=2.3cm]{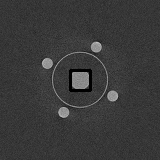}} & \adjustbox{valign=m}{\includegraphics[width=2.3cm]{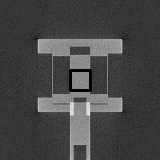}} \\
			\multirow{2}{*}{\rotatebox[origin=c]{90}{Circular}}  &  \adjustbox{valign=m}{\includegraphics[width=2.3cm]{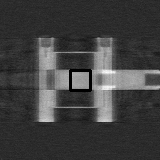}} & \adjustbox{valign=m}{\includegraphics[width=2.3cm]{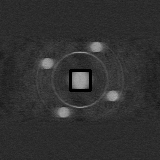}} & \adjustbox{valign=m}{\includegraphics[width=2.3cm]{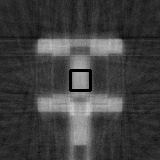}}  \\
			\multirow{2}{*}{\rotatebox[origin=c]{90}{GRU}}  &  \adjustbox{valign=m}{\includegraphics[width=2.3cm]{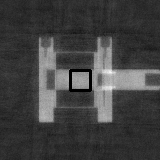}} & \adjustbox{valign=m}{\includegraphics[width=2.3cm]{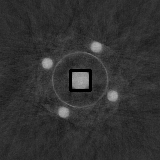}} & \adjustbox{valign=m}{\includegraphics[width=2.3cm]{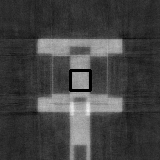}} 
		\end{tabular}
	\caption{Optimized CT trajectory for the center voxel. The optimized and circular CT trajectories consist of $k =50$ projections, while the reference data was a spherical CT trajectory with 1000 projections. All these images are reconstructed using the Algebraic Reconstruction Technique. The black square in each image represents the selected VOI. }
	\label{fig:complexCubeWorstCase}
\end{figure}

We reconstructed volume data from these 50 projections and compared them to reconstructions from a 50-projection circular trajectory and a 1000-projection spherical reference trajectory (Figure \ref{fig:complexCubeWorstCase}). The quantitative evaluation included the calculation of the Structural Similarity Index (SSIM), the Contrast-to-Noise Ratio (CNR) and the Peak Signal to Noise Ratio (PSNR). The GRU-optimized trajectory showed fewer artefacts and better image quality, as demonstrated by higher SSIM, PSNR, and CNR metrics compared to the circular trajectory (table \ref{tab:metrics_a}).

\begin{table}[htbp]
\centering
\caption{Comparison of Quality Metrics for 50 Projections}
\begin{tabular}{@{}ccccc@{}}
\toprule
\textbf{Approach} & \textbf{SSIM}$\uparrow$ & \textbf{PSNR}$\uparrow$ & \textbf{CNR}$\uparrow$  & \textbf{Coverage}   \\ 
\midrule
Circular & 0.38 & 120.0db  &  6.97 &45.6\%\\
GRU &  0.49 & 121.0db & 9.08 & 60.2\%\\ 
\bottomrule
\end{tabular}%
\label{tab:metrics_a}
\end{table}

\section{Conclusion and Outlook}

This study explores the use of GRU networks to optimize CT scan trajectories based on projection metrics. We first compute metrics for each projection of a test specimen, which are then fed into a GRU network to determine an optimized CT trajectory with 50 projections. The results show that this GRU-optimized trajectory outperforms a standard circular trajectory, highlighting the potential of GRUs in trajectory optimization. However, our results are based on a single specimen and simulated data, indicating the need for more extensive testing and application to real CT scan data.

Future research could also focus on the impact of different weights assigned to the projection-based metrics. Currently, we assign a weight of 16 to data completeness and 1 to other metrics. As indicated by our parameter search, different weights could significantly alter the optimized trajectories, warranting further investigation into metric weighting for improved optimization.

\end{document}